\documentclass[sigconf, nonacm]{acmart}
\newcommand\vldbyear{2025}
\newcommand\vldbworkshop{DaSH: Data Science with Human in the Loop}
\newcommand\vldbauthors{\authors}
\newcommand\vldbtitle{\shorttitle} 
\newcommand\vldbavailabilityurl{}
\newcommand\vldbpagestyle{plain} 

\usepackage{enumitem}
\usepackage{float}
\usepackage[most]{tcolorbox} 
\usepackage{xcolor}
\usepackage{listings}
\usepackage{xcolor} 
\usepackage{cleveref}
\usepackage{makecell}

\begin{document}
\title{Adobe Summit Concierge Evaluation with Human in the Loop}
%

\author{Yiru Chen$^\dagger$}
\affiliation{%
  \institution{Adobe Inc.}
}
\email{yiruc@adobe.com}

\author{Sally Fang$^\dagger$}
\affiliation{%
  \institution{Adobe Inc.}
}
\email{xinf@adobe.com}

\author{Sai Sree Harsha$^\dagger$}
\affiliation{%
  \institution{Adobe Inc.}
}
\email{ssree@adobe.com}

\author{Dan Luo$^\dagger$}
\affiliation{%
  \institution{Adobe Inc.}
}
\email{dluo@adobe.com}

\author{Vaishnavi Muppala$^\dagger$}
\affiliation{%
  \institution{Adobe Inc.}
}
\email{mvaishna@adobe.com}

\author{Fei Wu}
\affiliation{%
  \institution{Adobe Inc.}
}
\email{feiw@adobe.com}

\author{Shun Jiang}
\affiliation{%
  \institution{Adobe Inc.}
}
\email{shunj@adobe.com}

\author{Kun Qian}
\affiliation{%
  \institution{Adobe Inc.}
}
\email{kunq@adobe.com}

\author{Yunyao Li}
\affiliation{%
  \institution{Adobe Inc.}
}
\email{yunyaol@adobe.com}

\thanks{
$^{\dagger}$ Equal contribution. Authors are listed in alphabetical order.
}
\newcommand{\yiru}[1]{\noindent{\color{orange}{ #1}}}
\newcommand{\dan}[1]{\noindent{\color{red}{#1}}}
\newcommand{\sai}[1]{\noindent{\color{blue}{#1}}}
\newcommand{\shun}[1]{\noindent{\color{purple}{#1}}}
\newcommand{\vaishnavi}[1]{\noindent{\color{teal}{#1}}}
\newcommand{\kun}[1]{\noindent{\color{olive}{#1}}}
\newcommand{\SummitConcierge}{\textsc{Summit Concierge }}
\newcommand{\highlight}[1]{{\setlength{\fboxsep}{1pt}\colorbox{yellow}{\textbf{\texttt{#1}}}}}

\lstset{
	language = SQL,
	showspaces=false,
	basicstyle=\ttfamily\scriptsize,
	commentstyle=\color{gray},
	mathescape=true,
	numbers=none,
    escapeinside={^}{^},
	captionpos=b,	
	float=tp,
	floatplacement=tbp,
	belowskip=-0.05em,
    breaklines=true,
    frame=tlrb,
    xleftmargin=0pt,xrightmargin=0pt
} 

\crefname{listing}{Listing}{Listings}
\Crefname{listing}{Listing}{Listings}
\crefname{lstlisting}{Listing}{Listings}
\Crefname{lstlisting}{Listing}{Listings}
\crefname{figure}{Figure}{Figures}
\Crefname{figure}{Figure}{Figures}

\begin{abstract}
Generative AI assistants offer significant potential to enhance productivity, streamline information access, and improve user experience in enterprise contexts. In this work, we present \SummitConcierge, a domain-specific AI assistant developed for Adobe Summit. The assistant handles a wide range of event-related queries and operates under real-world constraints such as data sparsity, quality assurance, and rapid deployment. To address these challenges, we adopt a human-in-the-loop development workflow that combines prompt engineering, retrieval grounding, and lightweight human validation. We describe the system architecture, development process, and real-world deployment outcomes. Our experience shows that agile, feedback-driven development enables scalable and reliable AI assistants, even in cold-start scenarios.
\end{abstract}

\maketitle

\pagestyle{\vldbpagestyle}
\begingroup\small\noindent\raggedright\textbf{VLDB Workshop Reference Format:}\\
\vldbauthors. \vldbtitle. VLDB \vldbyear\ Workshop: \vldbworkshop.\\ 
\endgroup
\begingroup
\renewcommand\thefootnote{}\footnote{\noindent
This work is licensed under the Creative Commons BY-NC-ND 4.0 International License. Visit \url{https://creativecommons.org/licenses/by-nc-nd/4.0/} to view a copy of this license. For any use beyond those covered by this license, obtain permission by emailing \href{mailto:info@vldb.org}{info@vldb.org}. Copyright is held by the owner/author(s). Publication rights licensed to the VLDB Endowment. \\
\raggedright Proceedings of the VLDB Endowment. 
ISSN 2150-8097. \\
}\addtocounter{footnote}{-1}\endgroup

\ifdefempty{\vldbavailabilityurl}{}{
\vspace{.3cm}
\begingroup\small\noindent\raggedright\textbf{VLDB Workshop Artifact Availability:}\\
The source code, data, and/or other artifacts have been made available at \url{\vldbavailabilityurl}.
\endgroup
}

\section{Introduction}

Generative AI assistants offer enterprises tremendous potential, including substantial productivity gains, reducing barriers to entry, accelerating product adoption, amplifying creative workflows, and improving user experiences for both customers and employees~\cite{maharaj2024evaluation}. 
These systems can serve as intuitive, conversational interfaces to enterprise knowledge, streamlining access to information. 
However, developing a reliable, task-aligned assistant remains a complex and iterative process that must reconcile scalability with precision, user expectations with system capabilities, and time-to-market with development robustness.

\SummitConcierge is a generative AI assistant developed specifically for Adobe Summit~\footnote{\url{https://business.adobe.com/summit/adobe-summit.html}}, an annual event attracting thousands of attendees. The assistant is designed to handle a wide variety of event-related queries, ranging from session recommendations and speaker information to venue logistics and agenda search, delivering timely, accurate responses in natural language. By doing so, \SummitConcierge aims to enhance the attendee experience, reduce the burden on support staff, and provide scalable, real-time access to Adobe Summit information.
Despite the impressive capabilities of large language models~(LLMs), building such an assistant within a short amount of time presents several critical challenges:
1) \textbf{Data Sparsity}.
    The sparsity of historical query logs and interaction patterns makes it difficult to anticipate user intents or generate representative datasets for system training and evaluation. This cold-start problem limits the effectiveness of conventional fine-tuning and retrieval-based strategies that typically rely on real-world usage data.
2) \textbf{Reliable Quality}.  
    While LLMs excel at generating fluent and coherent responses, they often suffer from hallucinations or inaccuracies, especially when handling specific or time-sensitive information. Ensuring that the assistant consistently produces trustworthy and contextually grounded answers is essential for user trust and adoption.
3) \textbf{Quick to Deployment}.  
    Given the fixed timeline and one-time nature of Adobe Summit, the assistant must be developed, tested, and deployed rapidly. This constraint necessitates agile development practices, streamlined tooling, and minimal reliance on large-scale supervised training or complex engineering pipelines.

To address the challenges outlined above,  we adopt a \textit{human-in-the-loop} development paradigm, which brings human expertise into the loop to guide data curation, response validation, and quality monitoring. This hybrid approach enables rapid iteration and reliability without requiring extensive pre-collected data or long training cycles.

Our main contributions include:
1) \textbf{A human-in-the-loop workflow for quality assurance.}
We introduce a lightweight feedback loop involving human reviewers to continuously validate and refine responses, ensuring factual accuracy and contextual appropriateness in user interactions.
2) \textbf{Techniques to overcome data sparsity in cold-start scenarios.}  
    We describe how synthetic queries, documentation-grounded retrieval, and prompt engineering can effectively bootstrap an assistant in the absence of historical usage data.
3) \textbf{Lessons learned from real-world deployment at scale.}  
    We share empirical insights and design decisions from the production deployment of \SummitConcierge during Adobe Summit, including operational challenges, user feedback, and opportunities for future improvement.

\vspace{-10pt}
\section{Challenges}

Despite active research on AI assistance~\cite{sajja2024artificial, klemmer2024using}, building a reliable enterprise AI assistant quickly remains underexplored.

\vspace{-5pt}
\paragraph{Off-the-shelf benchmarks cannot work well for domain AI Assistants}
Although public benchmark datasets for general tasks are abundant (e.g., \citet{chang2024survey} lists 46 public benchmark datasets), they are often not applicable for domain-specific AI assistants. 
Creating domain-specific benchmark datasets is labor intensive, time-consuming, and requires domain expertise.
Moreover, assistants’ workload and tasks may also evolve. Thus, there is no one static benchmark data that suits all~\cite{danilevsky2020survey}. Therefore, benchmark data creation itself is a challenging task.

\vspace{-5pt}
\paragraph{Efficient iteration is needed}
Given the challenges of creating domain-specific benchmark datasets, it is essential to establish a process that supports rapid and efficient iteration. Unlike static datasets, evaluation benchmarks for domain AI assistants must evolve alongside product features, user needs, and task definitions~\cite{perez2022red}. 
This dynamic nature calls for lightweight mechanisms to continuously refine benchmarks—such as human-in-the-loop feedback~\cite{christiano2017deep}, prompt-based evaluations~\cite{zheng2023judging}, and LLM-as-judge protocols~\cite{li2024generation}. These methods allow teams to quickly assess assistant quality, identify failure cases, and update test scenarios without extensive manual labeling. 
Ultimately, efficient iteration enables faster alignment between assistant capabilities and real-world demands.

\vspace{-5pt}
\section{Summit Concierge Overview}

\begin{figure*}[h]
\centering
\includegraphics[width=0.93\linewidth]{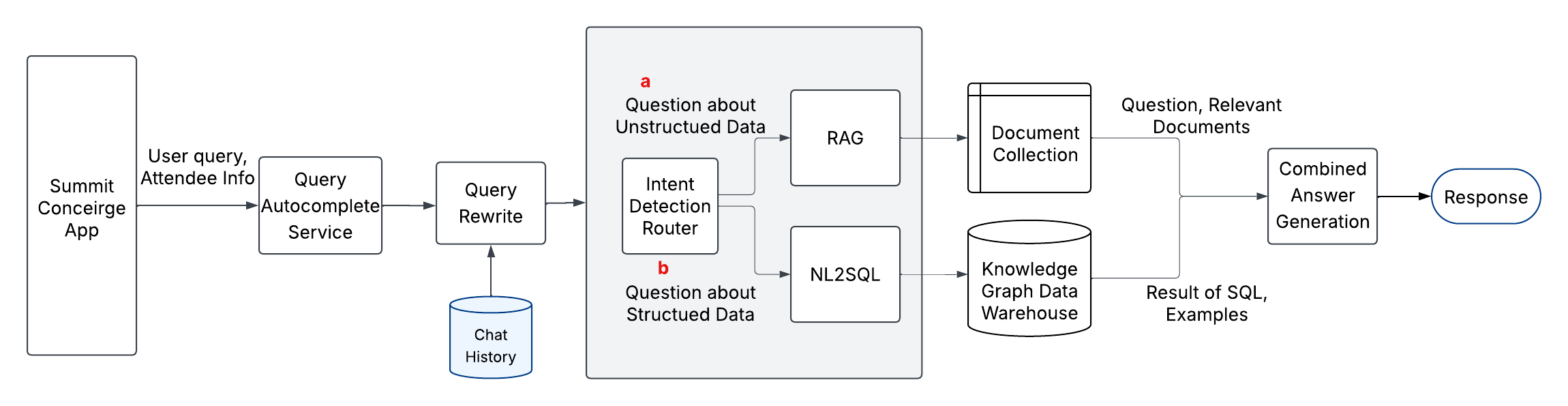}
\vspace{-10pt}
\caption{The overview of Summit concierge. }
\label{fig:sc_overview}
\vspace{-10pt}
\end{figure*}

In this section, we briefly introduce \SummitConcierge, an enterprise AI assistant designed to handle two primary types of user queries: 
(1) \textbf{General information queries}, which are answered using unstructured content from the Adobe Summit guidebook (e.g., \textit{What is included in the Summit Concierge?} or \textit{Where can I park during the Summit?}). In this paper, they are referred to as {Product Knowledge};
and 
(2) \textbf{Specific detail-oriented queries}, which require access to structured databases such as a knowledge graph or SQL warehouse 
(e.g., 
``\texttt{who is \{speaker\_name\}}'').
We refer to the guidebook as the \textit{unstructured data source}, and structured databases as the \textit{structured data source}. In this paper, they are referred to as {Operational Insight}.


An overview of the \SummitConcierge\ system architecture is illustrated in Figure~\ref{fig:sc_overview}. Following the standard LLM-based conversational question answering pipeline, the user query, combined with attendee-specific information, first passes through an autocomplete module and a query rewriting component that leverages chat history to improve clarity and completeness. 
The rewritten query is then routed by an intent detection module, which classifies it as either unstructured or structured. 
For unstructured queries, a retrieval-augmented generation (RAG) module retrieves relevant documents to ground the response. 
For structured queries, a natural language-to-SQL~(NL2SQL) module generates a query over the knowledge graph or data warehouse. The results from either branch are then passed to a unified answer generation module, which synthesizes the final response.
We elaborate the unstructured data source and structured data source in the following subsections.

\vspace{-10pt}
\subsection{Structured Data Source}
The structured data source contains information about the Summit collected and provided by the organizers through RainFocus, an event management tool.
It includes
1) Sessions information, including the title, abstract and logistic information such as time, location, duration and so on.
2) Speakers information, including the name, title, biography, company, and session they will be presenting at.
3) Sponsors information, including the company name, URL, level of sponsorship, and event booth location.
This data are ingested as relational database tables by a data transformation pipeline that runs periodically to ensure data is up-to-date.

In addition, schedules information for each individual attendee is also available, which includes all sessions the Summit
attendee has signed up for, and is particularly useful for answering personalization queries such as \textbf{\textit{``Where is my next session?''}}, and
 \textbf{\textit{``What session should I attend tomorrow about digital marketing''}}. Unlike the other information, attendee
schedule is private and more time sensitive. Information about sessions and speakers are mostly static during the Summit, while schedule information changes
as the attendees adding and removing sessions to their schedule using the Summit mobile app.  To ensure privacy and freshness, we did not ingest attendee information
as part of the Snowflake database, but instead, we use the RainFocus API to query the schedule information for each individual attendee when they ask a personalization query.
Given a query issued by users (e.g., ``\textbf{\textit{What session should I attend tomorrow about digital marketing}}''), we prompt the LLM to generate a SQL query to obtain
general information (in this case, 
``\texttt{all sessions about digital marketing on \{date\}}''). In parallel, we query RainFocus API with the appropriate attendee ID and authentication token
to obtain the schedule information for that attendee. We then render SQL execution result and attendee schedule as a tabular documentation in MarkDown format, and pass it to the LLM to generate a final
response.

\vspace{-10pt}
\subsection{Unstructured Data Source}
The unstructured data source used by Summit Concierge is composed of a corpus of rich, free-form textual content. The central component of this corpus is the ABC Guide, a comprehensive internal document curated by the event management team. It contains detailed information about the event logistics, agenda highlights, venue maps, FAQs, and operational protocols, originally intended to assist support staff but repurposed by the \SummitConcierge to address general attendee queries.
Key categories from the ABC Guide include:
1) Event logistics and navigation: agendas, venue maps, transportation, hotels, parking, rideshare locations, and registration processes.
2) Onsite services: food and beverage locations, coffee stations, first aid, accessibility services (e.g., ADA, wheelchair rentals), and internet access.
3) Attendee support and safety: information desks, coat/bag check, lost and found, security, health and safety guidance.

Additional unstructured content includes:
1) Adobe product summaries and new product announcements, which provide high-level overviews and key value propositions of features launched or highlighted during the event.
2) Live-authored content, generated dynamically during the event based on attendee interactions. This includes responses to frequently asked questions, clarifications on session content, and updates provided by the concierge team. Engineers monitored user feedback, chat logs, and search traffic to identify information gaps and quickly draft new content in collaboration with the event management team.

\vspace{-10pt}
\section{Question Generation}
In this section, we introduce our practice of human-in-the-loop development paradigm, which integrates human expertise throughout the \SummitConcierge development lifecycle. 

\vspace{-10pt}
\subsection{Structure Data Source Related Question Generation}



The structured dataset includes information about sessions, speakers, demos, and sponsors, which are the primary categories that attendees likely to ask about during the event. Since interacting with speakers, attending specific sessions, and exploring demos are among the main purposes of most attendees, we anticipate a high volume of user queries seeking precise information related to these categories. 
Our goal in generating questions based on the structure data is twofold: first, to create a high-quality evaluation set for measuring the performance of the AI Assistant; and second, to proactively predict common user queries, which enables us to build an autocomplete suggestion pool, allowing popular questions to be answered quickly and reliably. Given these objectives, the generated questions must satisfy key requirements: \textbf{it should be diverse, comprehensive, and representative of real-world user intents}. To construct such a dataset in low-resource setting, we adopted two human-in-the-loop approaches.

First, we work with product managers~(PMs) and the Adobe Summit marketing team to curate an initial set of 10 high-value seed questions, such as 
``\texttt{tell me more about \{speaker\_name\}}'', which are expected to be common during the event. These templated queries are then expanded using an LLM, prompted to generate paraphrases in a concise, smartphone-typing style—short, direct, and often using abbreviations or casual phrasing for faster input, as result, we are able to expand the seed templates to a much larger set of templates such as, 
``\texttt{who is \{speaker\_name\}}'', ``\texttt{tell me abt \{speaker\_name\}}'', which gives us more than 17K queries based on the initial seed templates from Subject Matter Expert~(SME).

Second, we use SQLSynth~\cite{tian2025text}, a human-in-the-loop tool that generates natural language questions from a given database schema. It first programmatically creates executable SQL queries and then derives corresponding questions. SQLSynth offers two modes: a fully automatic mode, where the entire process runs without intervention, and a human-in-the-loop mode, where users can review, revise, or reject the generated questions to improve fluency and naturalness. This reverse-engineering approach to text-to-SQL offers two key advantages:
1) \textbf{Scope Assurance}. By generating questions from executable SQL over the schema, all questions are guaranteed to be in-scope, unlike unconstrained LLM-generated questions, which may fall outside the schema and hinder evaluation.
2) \textbf{Diversity}. Programmatic SQL generation systematically explores varied attribute combinations, yielding a broader and more diverse set of natural language questions, including many that human annotators might overlook.

Without any human supervision, SQLSynth enables us to generate a large set of diverse, in-scope queries for stress testing the intent detection module, which routes valid questions to the structured data path. Since all generated queries are in-scope, the module should route them accordingly, and none should be rejected by NL2SQL. Using 269 auto-generated queries produced less than 30 minutes, we found the router achieved 100\% accuracy, while NL2SQL missed 2.6\%. Thanks to these these evaluation results, we then improved the prompt instruction to address some of the NL2SQL issues identified.

\vspace{-10pt}
\subsection{Unstructured Data Source Related Question Generation}
\label{sec:uns_question}

To fully utilize the breadth of unstructured content available in the ABC Guide and related sources, we develop a set of natural language questions targeting three core use cases: evaluation, autocomplete suggestions, and follow-up question generation. 
Given the informal and high-variance nature of these documents, we adopt LLM-assisted, human-in-the-loop strategies to ensure the resulting questions are diverse, high-quality, and aligned with real user behavior, especially in the mobile interaction context.

For evaluation, our goal is to construct a grounded question set tied directly to specific content sections within the BC Guide. 
We extract text passages and prompt a large language model~(LLM) to generate questions that are concise, representative of common user queries, and directly answerable based on the associated content.
Human reviewers then vet each question for fluency, naturalness, and answerability. 
The resulting set serves as an evaluation benchmark, where each question is linked to a source passage that provides a reliable gold-standard answer.


For follow-up question generation, we focus on extending conversations in a coherent and helpful manner. 
Given an existing question, its system-generated answer, and the supporting documents retrieved, we prompt the LLM to generate contextually relevant follow-up questions. 
These are subsequently reviewed for answerability and alignment with the prior exchange. Only those for which the system could produce high-confidence, accurate answers are retained. 
These are also stored in the question index and surfaced selectively during runtime to guide user engagement.

Across both use cases, the generation pipelines emphasize coverage, linguistic diversity, and practical answerability. 
The human-in-the-loop design ensures that model generations are carefully filtered and grounded, resulting in a reliable and representative dataset that enhances both evaluation rigor and end-user experience.

\vspace{-10pt}
\subsection{Autocomplete Question Generation}

For \SummitConcierge, attendees interact primarily through mobile devices; therefore, we specifically designed an autocomplete module to minimize typing effort and enhance overall usability and efficiency. 
As shown in Figure~\ref{fig:sc_overview}, the queries suggested by the autocomplete module are routed into the two aforementioned pipelines, depending on the detected intent. In this section, we particularly discuss our strategy to generate question evaluation for this module.

The questions for Autocomplete are generated through a human-in-the-loop process. 
The goal is to ensure both content diversity and popularity while maintaining contextual relevance, particularly in the case of multi-day events. To enhance the relevance of suggestions, event-specific questions are assigned start and expiration timestamps, enabling effective re-ranking based on their timeliness.
In the absence of historical data, two primary methods are employed to generate a list of seed questions. 
The first method involves directly curating questions from human experts and Large Language Models (LLMs), where we aim to proactively support users with frequently asked or high-value questions.
This set is constructed from three complementary sources. 
First, we apply a similar generation-and-review process as used for evaluation in Section~\ref{sec:uns_question}. 
Second, we incorporate curated questions from the event management team, based on historical data and anticipated needs. 
Third, we additionally log common queries received during the live event. 
All candidate questions are reviewed to ensure they elicited strong answers from the system. The validated subset is stored in a dedicated question index to support fast retrieval and consistent responses without invoking the LLM.


The second method involves defining question templates, which are subsequently instantiated into concrete questions by integrating contextual data. For example, a template like ``\texttt{Tell me more about <session\_name>}'',
can be instantiated into a specific query such as ``\textit{\textbf{Tell me more about the Opening Keynote session}}''. 
A rigorous quality control process is then applied to filter out questions that do not yield meaningful or relevant responses.

Once the seed questions are generated, they are incorporated into the Autocomplete function during internal bug bashes. Feedback and bug bash data provide insights into the coverage and popularity of the Autocomplete questions. Additionally, a human-in-the-loop clustering and popularity analysis method was employed to identify high-demand questions absent from the existing Autocomplete list. The approach combined semantic embeddings with HDBSCAN clustering to group similar queries, followed by manual review to filter noise and generalize representative question types. Clusters were ranked by frequency to surface the most commonly asked types. Human experts subsequently review and curate new questions and templates to ensure comprehensive coverage. 

By integrating human feedback in seed questions generation and the bug bash feedback review phases, both the overall quality of the Autocomplete question list and the effectiveness of popularity-based ranking have been significantly improved.

\section{Evaluation}

\subsection{Unstructured Data Evaluation}

We evaluate the quality of responses to questions grounded in unstructured data using three complementary strategies: correctness-based scoring, side-by-side comparison, and brand compliance checks. Each method captures a different aspect of response quality, and all use LLM judges with chain-of-thought reasoning calibrated for human alignment. Human reviewers are included in the loop for final validation in all three cases.
\vspace{-5pt}
\paragraph{Correctness Evaluation}
For questions with clear ground truth answers, we evaluate correctness by identifying a small set of key facts from the source content. Each response is decomposed into atomic claims, which are assessed for factual accuracy and groundedness. LLMs identify unsupported claims and missing key facts, and generate structured outputs for human review. We report the number of required facts covered, correct claims, and hallucinated content, offering fine-grained insight into factual reliability.

\vspace{-5pt}
\paragraph{Side-by-Side Comparison}
To support iterative improvement, we compare responses from different model variants or prompts. For each question, paired responses are judged for semantic equivalence. If differences are detected, the LLM explains them (e.g., factual inconsistency, verbosity, omission). This helps track regressions and improvements not captured by correctness alone. Disagreements are reviewed by humans to validate significant content differences.

\vspace{-5pt}
\paragraph{Brand Compliance Verification}
Ensuring brand alignment is essential in a public-facing assistant. We use an LLM-based checker to assess tone, language, and policy compliance against a set of guidelines provided by the event team. Violations are flagged with rationales using chain-of-thought reasoning. Human reviewers assess flagged outputs and make final decisions on compliance. This check is especially critical in production deployment.

This multi-faceted evaluation framework enables both high-precision benchmarking and ongoing refinement for unstructured-source response generation.
\vspace{-5pt}
\subsection{Templated Structured Data Query Evaluation}
\label{subsec: structeval}

Templated structured data queries are executed against the database to retrieve results, which are then represented to the user in natural language format.
For templated structured data query evaluation, since we have a limited number of predefined templates, we rely on data experts to write the corresponding gold SQL for each question. These gold SQL  are constructed using the same slot variables as the templated natural language queries. Executing the gold SQL gives us the key facts that must be included in the final natural language response.
For example, for a question template ``\texttt{What keynotes does <speaker> speak at?}'', the corresponding gold SQL is shown in \Cref{l:templated}. Both the template and the gold SQL share the same variable \highlight{<speaker>}.

Then, when given such a template and an instantiated query—What keynotes does Shantanu Narayen speak at?—we can infer that the value of the variable <speaker> is Shantanu Narayen. By substituting this value into the gold SQL shown in \Cref{l:templated}, we obtain a fully executable SQL query. Executing the query gives us key facts in a structured format with schema {\tt<title, speaker, start\_time, room>}, as shown in \Cref{l:keyfact}.

Given these gold key facts, we compare them with the response generated by the Summit Concierge agent to verify whether the key facts are present. We adopt a human-in-the-loop evaluation method that combines LLM judgment and human review. First, we use the LLM to determine whether the key facts in \Cref{l:keyfact} are covered in the response. The response is  shown in \Cref{fig:generated_response}. If the key facts match, the answer is marked as correct. If not, a human annotator further evaluates the case to decide whether the response is right. In this case, the answer in \Cref{fig:generated_response} is a correct answer.
Since we have a large number of queries, using the LLM as a first-pass judge accelerates the evaluation process, while human reviewers handle only the uncertain cases.

\begin{lstlisting}[language=SQL,
  caption = {The templated gold SQL for the question  -  What keynotes does \highlight{<speaker>} speak at?},
  label={l:templated}, breaklines=True]
SELECT DISTINCT session.title as title, to_char(session.start_time, 'YYYY-MM-DD %I:%M %p') as start_time, session.room as room
FROM session JOIN speaker ON session.session_id = speaker.session_id
WHERE speaker.full_name ILIKE ^\highlight{<speaker>}^ AND ARRAY_CONTAINS('Keynote'::variant, session.session_type) ORDER BY session.priority ASC
\end{lstlisting}

\begin{lstlisting}[
  caption = {The key facts for the query   -  What keynotes does Shantanu Narayen speak at?},
  label={l:keyfact}, 
  breaklines=True]
[title: Opening Keynote, start_time: 2025-03-18 09:00AM, room: Keynote Hall], 
[title: Day Two Keynote, start_time: 2025-03-19 10:00AM, room: Keynote Hall]
\end{lstlisting}

\paragraph{Evaluation Metrics}
During the Summit event, this LLM-as-judge approach helps us significantly reduce annotation effort. Out of 3,000 templated queries, only 220 requires human reviews. The rest are automatically verified as containing the necessary key facts.

\begin{figure}[h]
    \centering
    \begin{tcolorbox}[
        title=\texttt{What keynotes does Shantanu Narayen speak at?},
        width=0.95\columnwidth 
    ]
    \begin{flushleft}
\href{ebadobesummit:///index?original_serial=1710768905259001aR3y}{Shantanu Narayen} will be speaking at the following keynotes during Adobe Summit 2025:

\textbf{\href{ebadobesummit:///event?original_serial=1736451342668001sNkk}{1. Opening Keynote}}\\
\textbf{Time:} March 18, 2025, at 9:00 AM\\
\textbf{Room:} Keynote Hall\\
\textbf{Other Speakers:} \href{ebadobesummit:///index?original_serial=1596731161054001K3z8}{Jennifer Oleksiw}, 
\href{ebadobesummit:///index?original_serial=15976863006580015o3P}{Wes Hopkins}, 
\href{ebadobesummit:///index?original_serial=1732131011084001plbS}{James Quincey}, 
\href{ebadobesummit:///index?original_serial=1738955913618001w2JY}{Lara Balazs}, 
\href{ebadobesummit:///index?original_serial=1652393052924001kokL}{Anil Chakravarthy}, 
\href{ebadobesummit:///index?original_serial=16565305300090016iJf}{David Wadhwani}, 
\href{ebadobesummit:///index?original_serial=1603144507816001mWia}{Ann Rich}

\vspace{1em}

\textbf{\href{ebadobesummit:///event?original_serial=1736451477872001VEZt}{2. Day Two Keynote}}\\
\textbf{Time:} March 19, 2025, at 10:00 AM\\
\textbf{Room:} Keynote Hall\\
\textbf{Other Speakers:} \href{ebadobesummit:///index?original_serial=1733862607473001v1rO}{Jamie Dimon}, 
\href{ebadobesummit:///index?original_serial=1549575506787001hmbP}{Amy James}, 
\href{ebadobesummit:///index?original_serial=1736973515362001i3On}{Esi Eggleston Bracey}, 
\href{ebadobesummit:///index?original_serial=1677102999256001VyfG}{Hilary Cook}, 
\href{ebadobesummit:///index?original_serial=1549411708991001HZe2}{Amit Ahuja}, 
\href{ebadobesummit:///index?original_serial=1549658845030001hK32}{Anjul Bhambhri}, 
\href{ebadobesummit:///index?original_serial=1733235838325001h1RR}{Rachel Thornton}, 
\href{ebadobesummit:///index?original_serial=1652393052924001kokL}{Anil Chakravarthy}, 
\href{ebadobesummit:///index?original_serial=1619456329094001kNCu}{Rachel Hanessian}, 
\href{ebadobesummit:///index?original_serial=1718305424403001f0Nr}{Colin Fleming}

\vspace{1em}

These sessions promise to be insightful and engaging!

\href{ebadobesummit:///sessions?category_ids=2124009296}{Tap here to explore Keynote sessions}    \end{flushleft}
    \end{tcolorbox}
\vspace{-10pt}
    \caption{\SummitConcierge  generated response for the query - "What keynotes does Shantanu Narayen speak at?". 
    \vspace{-10pt}
}
\label{fig:generated_response}   
\end{figure}



\subsection{Autocomplete Question Pool Evaluation}
The evaluation of our autocomplete system is based on how effectively the question pool can suggest relevant prompt completions to enhance the user experience. Our evaluation protocol consists of two components:
First, we adopt an LLM-as-judge framework to assess the quality of the autocomplete suggestions. This allows us to quantify the number of queries for which the system provides relevant and helpful completions.
Second, we measure the reduction in user typing effort by evaluating the number of keystrokes saved. Specifically, for each input query, we generate all of its prefix expansions. For example, for the query ``\textbf{Tell me more about the Opening Keynote session}'', we construct prefixes such as ``T'', ``Te'', ``Tel'', and so on. For each prefix variant, we also adopt an LLM-as-judge framework to assess the quality of the autocomplete suggestions against the prefix variant. 
The number of keystrokes saved is then calculated as the length of the original query minus the minimum number of characters typed before a relevant suggestion appears among the top-ranked completions.

\vspace{-5pt}
\paragraph{Evaluation Metrics}
Our evaluation results across three different question pools are presented in Table~\ref{tab:autocomplete_metrics}. These question pools are developed in chronological order. We observe that, by adopting our proposed approach, the quality of autocomplete suggestions consistently improves across the iterations.

\begin{table}[h]
\small
\setlength{\tabcolsep}{2.5pt} 
\centering
\vspace{-10pt}
\caption{Evaluation  of autocomplete question pools.}
\vspace{-10pt}
\label{tab:autocomplete_metrics}
\begin{tabular}{p{2.5cm}p{2.5cm}p{2.5cm}}
\toprule
\textbf{Question Pool} & \textbf{Ratio of Relevant Completion} & \textbf{Avg. Keystroke Saving} \\
\midrule
Question Pool\_1 & 27.00\% & 6.09 \\
Question Pool\_2 & 48.85\% & 8.85 \\
Question Pool\_3 & 58.02\% & 11.45 \\
\bottomrule
\end{tabular}
\vspace{-20pt}
\end{table}

\subsection{Multi-turn Evaluation}
Except for the evaluation above, the user experience is also critical. 
In the \SummitConcierge, users engage in multi-turn dialogues that reference previous conversational context and contain temporal ambiguity, since the event spans multiple days. Users may type shorter or more ambiguous sentences (e.g., ``\textbf{What sessions are after lunch?}'' or simply ``\textbf{Tomorrow?}''), relying on the system to infer the intended time frame or context from the prior conversation. 
Accurately resolving such queries requires contextual and temporal understanding to support accurate Retrieval-Augmented Generation~(RAG). To address this, we use a reasoning-oriented Large Language Model~(LLM) for prompt rewriting, using chain-of-thought prompting and few-shot examples with step-by-step reasoning. This approach enables the model to explicitly analyze the user’s input, resolve ambiguities, and produce clear, contextually grounded rewrites; for instance, transforming an ambiguous query like 
``\textbf{What sessions are available after lunch?}'' into 
``\textbf{What sessions are scheduled after 1 PM on Day 2 of the conference?}'' 

For evaluation, we utilize both automated and human-in-the-loop methods. The LLM serves as a judge to assess factual correctness and identify any remaining ambiguities in both the original and rewritten queries, establishing effective ground-truth labels. We further enhance evaluation robustness by incorporating human annotation, especially for sampled out-of-scope (OOS) questions and cases where rewrites might incorrectly alter user intent or impact downstream tasks. We review these instances to ensure that the chain of reasoning in prompt rewriting maintains user intent and appropriately addresses context. Additionally, we continuously monitor user feedback from the deployed system, rapidly flagging and annotating incorrect rewrites identified by users or through system performance monitoring. This ongoing human-in-the-loop process, informed by real-world feedback and error analysis, allows us to continuously improve the system’s ability to handle ambiguous or under specified queries in multi-turn conversations.

\vspace{-5pt}
\paragraph{Evaluation Metrics}
We used LLM-as-a-Judge to auto select uncertain samples for the human review. The model classified outputs as rewrite correctness along with the confidence scores and key facts. This approach reduced manual annotation needs from 1500 queries to just 276, with the remaining auto-verified. The results before and after the prompt improvements are shown in Table~\ref{tab:mt_metrics}.

\vspace{-7pt}
\begin{table}[h]
\centering
\caption{Multi Turn Evaluation Metrics}
\vspace{-10pt}
\label{tab:mt_metrics}
\begin{tabular}{lrr}
\toprule
\textbf{Metric} & \textbf{Before} & \textbf{After} \\
\midrule
Rewrite Error Rate & 4.35\% & 1.45\% \\
Routing Accuracy & 89.1\% & 96.1\% \\
\bottomrule
\end{tabular}
\vspace{-12pt}
\end{table}

\section{Summit Outcomes}
As part of the assistant’s pre-event evaluation, a structured internal annotation effort was conducted over several days leading up to and during the event. Internal testers submitted queries across realistic structured and unstructured scenarios, with thumbs-down feedback and out-of-scope (OOS) flags guiding error analysis. Two rounds of human-in-the-loop review were performed daily by subject matter experts. In total, 624 interactions were triaged, surfacing recurring error themes, confirming prior fixes, and providing targeted insights for improvement efforts. 

\vspace{-5pt}
\subsection{Annotation Volume by Day}

Table~\ref{tab:daily_volume} shows the number of interactions annotated each day across the four-day internal evaluation period. This includes both pre-event triage and daily retrospectives conducted during the event. A consistent annotation cadence—two rounds per day—enabled the team to rapidly identify and resolve issues throughout the week.

\begin{table}[h]
\centering
\caption{Total Annotated Interactions by Day}
\vspace{-5pt}
\label{tab:daily_volume}
\begin{tabular}{lr}
\toprule
\textbf{Day} & \textbf{Annotated Interactions} \\
\midrule
Pre-Event / Day 1 Retro & 174 \\
Day 2 Retro             & 245 \\
Day 3 Retro             & 123 \\
Day 4 Retro             & 82  \\
\midrule
\textbf{Total}          & \textbf{624} \\
\bottomrule
\end{tabular}
\vspace{-5pt}
\end{table}

\vspace{-3pt}
\subsection{Annotation Summary}

Of 624 annotated interactions, 436 (69.9\%) did not require further action at the time they were reviewed. This group included queries that had already been addressed in earlier triage rounds, confirmed valid responses, or questions determined to be unsupported (out-of-scope). The remaining 188 interactions surfaced actionable issues across assistant components, including product knowledge errors, improvements to canned responses, rewrite errors, and intent routing issues.
Table~\ref{tab:annotation_summary} summarizes the outcomes of all 624 annotated interactions. Most were resolved or deemed valid or unsupported, while the remaining errors highlighted areas for targeted system improvements. As a result, the rate of queries incorrectly routed as out-of-scope (OOS) was reduced from 4\% to 3\%.

\vspace{-5pt}
\begin{table}[h]
\small
\centering
\caption{Summary of Annotation Outcomes (N = 624).}
\vspace{-5pt}
\label{tab:annotation_summary}
\begin{tabular}{lrr}
\toprule
\textbf{Category} & \textbf{Count} & \textbf{Percentage} \\
\midrule
\multicolumn{3}{l}{\textit{Resolved / No Issue}} \\

\quad \makecell[l]{Already Resolved, Not an Error,\\ or Out-of-Scope} & 436 & 69.9\% \\
\addlinespace
\multicolumn{3}{l}{\textit{Actionable Errors Addressed}} \\
\quad Product Knowledge Error               & 79  & 12.7\% \\
\quad Canned Response (Needs Improvement) & 31 & 5.0\% \\
\quad Rewrite Error                         & 21  & 3.4\% \\
\quad Intent Detection Error                & 20  & 3.2\% \\
\quad Operational Insights Error            & 30  & 4.8\% \\
\quad Tone of Response                      & 5   & 0.8\% \\
\quad Hyperlinking Issue                    & 2   & 0.3\% \\
\midrule
\textbf{Total} & \textbf{624} & \textbf{100\%} \\
\bottomrule
\end{tabular}
\vspace{-5pt}
\end{table}

\vspace{-5pt}
\section{Conclusion}

In this paper, we presented our experience developing \SummitConcierge, a generative AI assistant tailored for Adobe Summit. 
%
To address the data sparsity, ensuring response reliability, and deploying under strict timelines, we adopted a human-in-the-loop development paradigm that enabled rapid iteration and quality control without requiring extensive historical interaction data. Through prompt engineering, documentation-aware retrieval, and synthetic data augmentation, we were able to bootstrap the assistant and adapt it to the dynamic needs of Summit attendees. Our real-world deployment demonstrated the practical benefits of combining scalable LLM capabilities with lightweight human oversight, resulting in improved user experience and reduced operational overhead.

Looking forward, we see opportunities to generalize our methodology to other enterprise domains, particularly those that involve event support, internal knowledge access, or customer service. We believe that integrating agile human feedback loops with prompt-centric development and retrieval-based grounding offers a viable path to reliable, domain-specific AI assistants at scale.
\section{Acknowledgment}
We would like to thank Akash Maharaj, Kai Lau, Saurabh Stripathy, Jeremy Shi, and Brain Shin for their valuable contributions to the annotation and evaluation of the work presented in this paper. Their support and insights were instrumental in the development and validation of our system.

\bibliographystyle{ACM-Reference-Format}
\bibliography{dash}

\end{document}